\documentclass{ieeeaccess}
\usepackage{cite}
\usepackage{amsmath,amssymb,amsfonts}
\usepackage{algorithmic}
\usepackage{graphicx}
\usepackage{textcomp}
\usepackage{booktabs}
\usepackage{multirow} 

\usepackage{algorithm}
\usepackage[bookmarks=false,hidelinks]{hyperref}

\def\BibTeX{{\rm B\kern-.05em{\sc i\kern-.025em b}\kern-.08em
    T\kern-.1667em\lower.7ex\hbox{E}\kern-.125emX}}

\begin{document}
\history{Author accepted manuscript. Published in IEEE Access, vol.~14, pp.~172333--172348, 2026.}
\doi{\href{https://doi.org/10.1109/ACCESS.2026.3733819}{10.1109/ACCESS.2026.3733819}}

\title{Task-Aware Hybrid QUBO Optimization for Structured Neural Network Pruning}

\author{\uppercase{Osama Orabi}\authorrefmark{1,2,4}, 
\uppercase{Artur Zagitov}\authorrefmark{1}, 
\uppercase{Hadi Salloum}\authorrefmark{1,2,3}, 
\uppercase{Viktor A. Lobachev}\authorrefmark{4},
\uppercase{Kasymkhan Khubiev}\authorrefmark{5}, and
\uppercase{Yaroslav Kholodov}\authorrefmark{1,4,5}}

\address[1]{Laboratory of Quantum Computing, Innopolis University, 420500 Innopolis, Russia (e-mail: o.orabi@innopolis.university; a.zagitov@innopolis.university; ya.kholodov@innopolis.ru)}
\address[2]{Q Deep, Laboratory of Quantum Computing, 420502 Innopolis, Russia}
\address[3]{Research Center of the Artificial Intelligence Institute, Innopolis University, 420500 Innopolis, Russia (e-mail: h.salloum@innopolis.university)}
\address[4]{Moscow Independent Research Institute of Artificial Intelligence (MIRAI), 117218 Moscow, Russia (e-mail: VLOB1969@yahoo.com)}
\address[5]{Sirius University of Science and Technology, 354340 Sochi, Russia (e-mail: hubiev.k@talantiuspeh.ru)}

\markboth
{Orabi \headeretal: Task-Aware Hybrid QUBO Optimization for Structured Neural Network Pruning}
{Orabi \headeretal: Task-Aware Hybrid QUBO Optimization for Structured Neural Network Pruning}

\corresp{Corresponding author: Osama Orabi (e-mail: o.orabi@innopolis.university).}

\begin{abstract}
Neural network pruning can be formulated as a combinatorial optimization problem, yet many existing approaches rely on independent filter-importance scores or simplified objective functions. In this work, we propose a Hybrid Quadratic Unconstrained Binary Optimization (QUBO) framework for structured filter pruning that combines task-aware sensitivity information with interactions between candidate filters. The formulation incorporates first-order Taylor sensitivity and Weight-Fisher sensitivity into the linear component of the objective and can additionally incorporate activation similarity into the quadratic interactions. To control the target pruning cardinality without introducing an explicit quadratic cardinality penalty, we use a binary search over the capacity incentive to identify a coefficient that empirically yields the target pruning cardinality. We further investigate a two-stage QUBO--Tensor-Train refinement strategy in which the QUBO solution initializes a gradient-free probabilistic optimizer to search for improved pruning masks using the downstream metric. Experiments on the SIDD image denoising task and a Half-UNet model show that the Hybrid QUBO achieves higher PSNR and SSIM than the evaluated Taylor and L1-based QUBO baselines at the studied pruning target. Multi-seed experiments under a fixed dataset protocol are used to assess robustness, while controlled sub-problem experiments demonstrate that Tensor-Train refinement becomes increasingly valuable as the combinatorial problem size grows. The results support Hybrid QUBO as a task-aware structured pruning framework for the evaluated setting, while also highlighting the computational and deployment limitations of mask-based pruning.
\end{abstract}


\titlepgskip=-21pt

\maketitle
\pagestyle{plain}
\thispagestyle{plain}

% --- Introduction ---
\section{Introduction}
\label{sec:introduction}
\PARstart{M}{odern} deep Convolutional Neural Networks (CNNs) and large foundational models achieve exceptional accuracy across a wide range of complex domains. These applications span from standard image restoration to highly resource-intensive medical tasks, such as enhanced 3D brain tumor segmentation \cite{chaieb20263d} and multimodal neurodegenerative disease analysis \cite{chaieb2026brainmind}. However, their computational and memory requirements can limit deployment on resource-constrained devices \cite{liu2017learning,xu2024device,howard2017mobilenets,zhang2018shufflenet,han2015deep}.

Structured pruning—the removal of entire filters or channels—has become a standard solution to reduce FLOPs and memory footprint \cite{filters2016pruning,cheng2024survey,wen2016learning,he2017channel}. Unlike unstructured sparsity, structured removal can translate into practical reductions in computation, memory, and latency when the resulting architecture is physically reconstructed and efficiently executed on the target hardware. However, identifying the optimal subset of filters to retain under a strict parameter budget is fundamentally an NP-hard combinatorial optimization problem.

Quantum annealing has also been investigated as a computational paradigm for machine-learning optimization, with recent experimental studies examining its performance on real-world datasets and highlighting both its potential and practical limitations \cite{salloum2024integration,salloum2025performance}. In the context of neural network compression, recent works have begun treating pruning as an explicit combinatorial optimization task. Quadratic Unconstrained Binary Optimization (QUBO) formulations have been proposed to mathematically capture pairwise filter interactions \cite{wang2026quantum}. Unfortunately, the evaluated L1-based QUBO formulation relies primarily on weight magnitude, which can provide less task-specific information than gradient-based sensitivity criteria \cite{wang2026quantum}. Furthermore, a common way to encode hard sparsity constraints in QUBO is to introduce a quadratic cardinality penalty. Such penalties introduce additional quadratic terms and can substantially alter the energy landscape, creating steep ``walls'' that trap solvers in suboptimal local minima.

To manage this complexity, early pruning methods relied on greedy heuristics, ranking filters by weight-magnitude or activation statistics. First-order Taylor expansion of the loss function has since emerged as a widely used, task-aware criterion \cite{molchanov2019importance,theis2018faster}. Despite their efficiency, the basic scalar-score formulation does not explicitly model inter-filter interactions. To better capture data-driven sensitivity, some approaches utilize second-moment Fisher-information or Hessian-derived metrics—tracing back to foundational works like Optimal Brain Damage and Optimal Brain Surgeon, alongside modern trace approximations \cite{lecun1989optimal,hassibi1992second,navarretescalable,singh2020woodfisher,meyer2021hutch}. These cues complement first-order metrics by capturing the second moment of the gradient signal across the data distribution. Yet, functional redundancy remains a known issue that simplistic scoring misses; two filters may individually possess high importance scores while computing nearly identical features \cite{he2019filter,singh2020leveraging,shaikh2025dynamic}. 

In this work, we address these fundamental flaws by proposing a unified Hybrid QUBO framework. We formulate structured filter pruning as a global combinatorial formulation. Our formulation integrates gradient-aware sensitivity metrics (both first-order Taylor and Weight-Fisher sensitivity) into the linear term, and data-driven activation similarity into the quadratic term \cite{molchanov2019importance,liu2021group,geng2022pruning}. To circumvent the traditional pitfalls of hard penalty constraints, we introduce a \textbf{Dynamic Capacity Search} procedure that uses a binary search over the capacity incentive to identify a coefficient that empirically yields the target pruning cardinality without introducing an explicit quadratic cardinality penalty directly into the QUBO. 

Finally, because the QUBO objective provides a quadratic surrogate for the downstream pruning quality, we introduce a two-stage refinement process. Gradient-free optimizers have recently been used to refine discrete masks \cite{oseledets2011tensor,cichocki2017tensor}. PROTES has been proposed for high-dimensional discrete optimization using a Tensor-Train parameterization of the search distribution \cite{batsheva2023protes}. In our framework, it is used as a derivative-free refinement stage initialized from the QUBO solution. Its benefit is therefore an empirical question that depends on the dimensionality of the pruning problem and the available evaluation budget.

Our contributions are as follows:
\begin{enumerate}
    \item We propose a \textbf{Hybrid QUBO} objective that combines task-aware importance information with inter-filter redundancy modeling to search the discrete solution space.
    \item We integrate first-order Taylor sensitivity and Weight-Fisher sensitivity into the QUBO objective, providing complementary local and second-moment information for task-aware filter selection.
    \item We introduce a \textbf{Dynamic Capacity Search} procedure that uses a binary search over the capacity incentive to identify a coefficient that empirically yields the target pruning cardinality.
    \item We investigate a two-stage QUBO--Tensor-Train refinement strategy in which the QUBO solution provides an initialization for subsequent gradient-free probabilistic optimization to search for improved pruning masks using the downstream metric.
    \item We provide multi-seed statistical evaluation under a fixed dataset split, together with paired statistical tests and effect-size analysis, to assess the robustness of the principal pruning configurations.
\end{enumerate}

\paragraph{Paper organization.}
The remainder of this paper is organized as follows. Section II reviews structured pruning, optimization-based pruning, QUBO formulations, and tensor-based derivative-free optimization. Section III presents the proposed pruning formulation, including the binary pruning variables, importance measures, redundancy modeling, Fisher sensitivity, normalization, and dynamic capacity search. Section IV describes the dataset, model, baselines, experimental protocol, and reproducibility settings. Section V reports the pruning results, statistical robustness analysis, ablation studies, dynamic capacity search validation, Tensor-Train scaling, cross-task evaluation, and computational cost. Section VI discusses the performance--cost trade-off, limitations, and practical implications of the proposed framework. Section VII concludes the paper.

% --- Related Work ---
\section{Related Work}
\label{sec:related_work}
Structured pruning has become an important direction in neural network compression because removing complete filters or channels can produce models that are more compatible with conventional hardware than unstructured sparsity. Recent surveys organize structured pruning methods according to pruning granularity, timing, optimization strategy, and whether the method provides architecture- or hardware-aware acceleration \cite{cheng2024survey}.

\subsection{Magnitude-, Regularization-, and Gradient-Based Pruning}
Early structured pruning methods commonly ranked filters according to weight magnitude or other inexpensive statistics. Network Slimming instead introduces sparsity-inducing regularization on channel scaling factors during training \cite{liu2017learning}. Taylor-based methods subsequently introduced task-aware first-order sensitivity criteria, allowing filters to be ranked according to their estimated contribution to the task loss \cite{molchanov2019importance,theis2018faster}. These methods are computationally attractive, but they generally make pruning decisions using scalar filter scores and therefore do not explicitly model interactions between filters.

\subsection{Redundancy- and Activation-Aware Pruning}
A complementary line of work focuses on redundancy between filters. FPGM uses geometric-medoid/geometric-median information to identify filters that are redundant in weight space \cite{he2019filter}. ThiNet formulates filter selection using statistics obtained from subsequent layers, explicitly accounting for the effect of a retained subset on downstream computations \cite{luo2017thinet}. Other approaches combine magnitude or importance scores with correlation-based redundancy measures \cite{singh2020leveraging,geng2022pruning,wang2021convolutional,shaikh2025dynamic}. 

Our approach follows the same motivation of jointly considering candidate filters, but encodes pairwise score interactions directly into a quadratic optimization problem. Functional redundancy is additionally modeled through activation similarity in the extended formulation.

\subsection{Optimization-Based and Search-Based Pruning}
Treating pruning as an explicit optimization problem has led to approaches based on reinforcement learning, evolutionary search, and mathematical programming. Reinforcement-learning approaches such as AMC learn layer-wise compression policies by exploring the trade-off between accuracy, computation, and model size \cite{he2018amc}, while combinatorial approaches formulate pruning decisions directly as discrete optimization problems.

Recent optimization-based pruning methods such as CHITA formulate pruning through local quadratic approximations of the loss and solve structured combinatorial subproblems under explicit sparsity constraints \cite{benbaki2023fast}. These approaches demonstrate the potential of optimization, but they address a different objective and parameterization from the QUBO formulation considered here.

\subsection{QUBO-Based Pruning}
QUBO provides a natural binary representation of structured pruning because each candidate filter can be represented by a binary decision variable. Our work builds specifically on QUBO-based structured pruning \cite{wang2026quantum}. Related QUBO-based approaches have also been investigated for neural-network quantization, where discrete compression decisions are formulated as combinatorial optimization problems \cite{orabi2026task}. In contrast, the present work focuses on structured filter pruning and explicitly incorporates task-aware Taylor sensitivity and Weight-Fisher sensitivity into the QUBO diagonal, and can model functional redundancy through activation similarity in the quadratic interactions. We additionally introduce a dynamic capacity-search procedure for achieving a prescribed pruning cardinality and investigate Tensor-Train refinement of the resulting discrete solution.

Thus, the contribution of this work is not the use of QUBO alone, but the construction of a task-aware QUBO objective that combines local sensitivity, second-moment sensitivity, pairwise score-interactions, and capacity control within a unified discrete optimization framework.

\subsection{Tensor-Based Derivative-Free Optimization}
Derivative-free optimization provides another route for searching for improved discrete pruning configurations after an initial solution has been obtained. Prior work has investigated Tensor-Train sampling for derivative-free optimization of QUBO-based machine-learning models, demonstrating the potential of TT methods for searching high-dimensional discrete solution spaces \cite{salloum2025black}. Tensor-Train approaches have also been investigated for QUBO-based computer-vision optimization, including unsupervised image segmentation \cite{salloumtensor}. 

PROTES introduces probabilistic optimization using Tensor-Train representations, allowing high-dimensional discrete search spaces to be explored through a compact parameterization \cite{batsheva2023protes}. This makes TT-based optimization potentially useful when the true evaluation metric is expensive and the objective landscape is poorly represented by a quadratic surrogate.

The present work therefore combines the two paradigms: QUBO provides an efficient task-aware combinatorial initialization, while PROTES is used as a downstream derivative-free refinement mechanism. The experimental analysis specifically investigates how the usefulness of this refinement changes with the dimensionality of the controlled pruning problem.

% --- Methodology ---
\section{Methodology}
\label{sec:methodology}

This section describes the complete pruning framework. We first define the pruning variables and optimization convention, then introduce the L1-based QUBO baseline and progressively incorporate Taylor sensitivity, Weight-Fisher sensitivity, activation similarity, and dynamic capacity control. Finally, we describe the Tensor-Train refinement stage used to directly improve the downstream metric beyond the QUBO proxy.

\begin{figure*}[t]
\centering
\includegraphics[width=\textwidth]{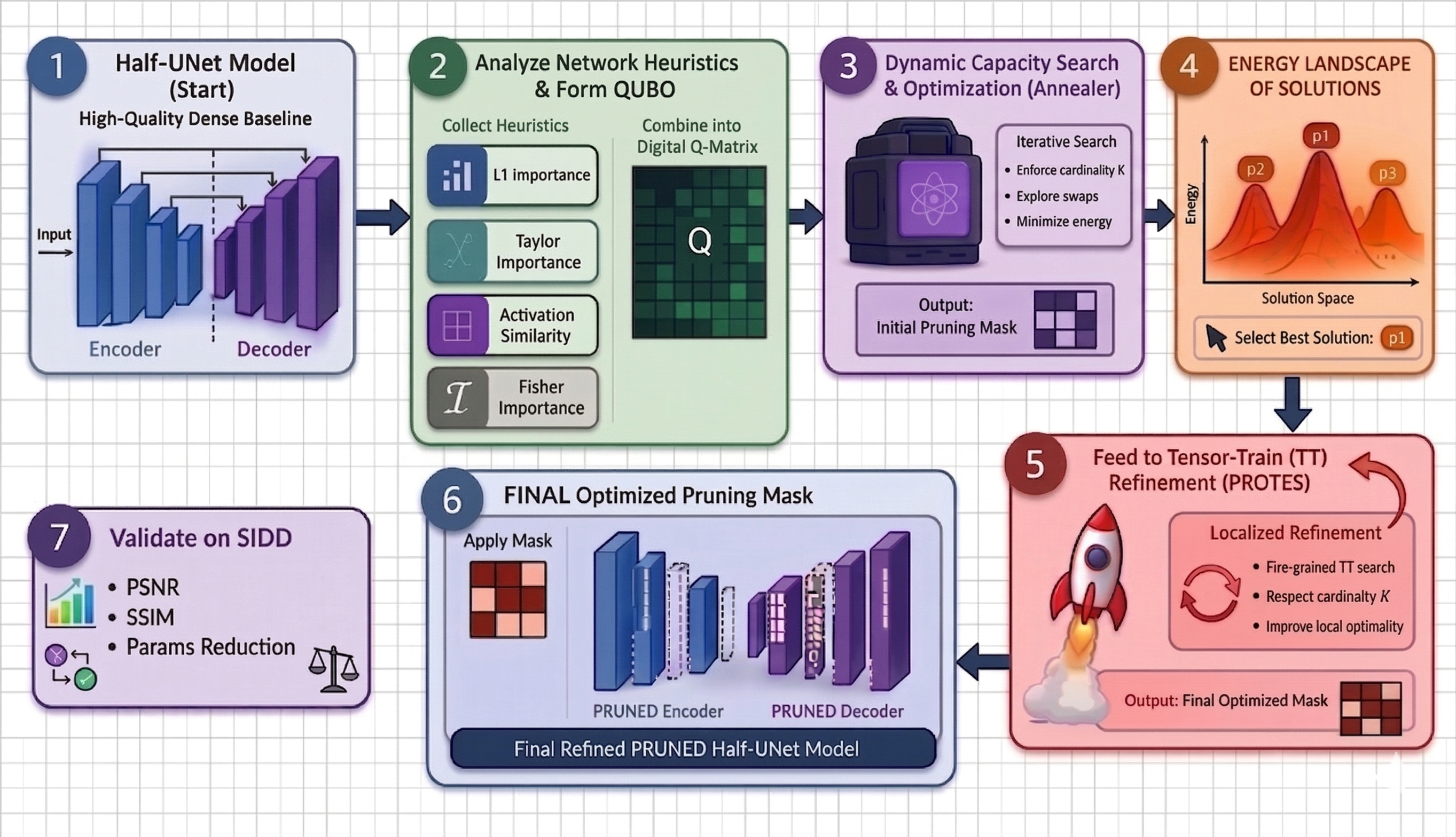}
\caption{Overview of the proposed hybrid pruning pipeline. The process begins by analyzing a dense baseline model to compute gradient sensitivities and activation redundancy (Stages 1-2). These heuristics populate a Hybrid QUBO matrix, which is searched using a dynamic capacity procedure (Stage 3) to navigate the discrete energy landscape (Stage 4). The highest-quality mask seeds a Tensor-Train (PROTES) refinement stage (Stage 5) to directly improve the true non-differentiable validation metric, yielding the final optimized network (Stages 6-7).}
\label{fig:pipeline}
\end{figure*}

Figure \ref{fig:pipeline} illustrates the complete workflow of our proposed two-stage pruning framework. The process transitions from initial heuristic analysis to global combinatorial formulation, and concludes with a probabilistic gradient-free tensor-train refinement to optimize the true downstream task performance.

\subsection{Baseline: Greedy Taylor pruning}
A well-established importance measure for structured pruning is the first-order Taylor expansion \cite{molchanov2019importance,theis2018faster}.
For filter parameters $W_i$, the Taylor importance is:
\begin{equation}
I_i^{\text{Taylor}} \;=\; \Big\lvert \frac{\partial \mathcal{L}}{\partial W_i} \odot W_i \Big\rvert_1,
\end{equation}
where $\mathcal{L}$ is the task loss and $\odot$ denotes elementwise multiplication.
Filters with the lowest $I_i^{\text{Taylor}}$ are greedily pruned. 
This baseline is computationally efficient and task-aware, but it ignores global interactions between filters and redundancy effects across layers.

\subsection{Binary pruning variables and optimization convention}
\label{subsec:binary_convention}

We define a binary pruning vector $\mathbf{p} \in \{0,1\}^N$ over $N$ prunable filters, where
\begin{equation}
p_i =
\begin{cases}
1, & \text{if filter } i \text{ is pruned}, \\
0, & \text{if filter } i \text{ is retained}.
\end{cases}
\end{equation}
All QUBO formulations in this work are posed as \emph{minimization} problems: lower energy corresponds to a better pruning configuration.
Under this convention, linear terms with positive coefficients discourage pruning, while negative linear terms encourage pruning.
This choice allows sparsity incentives to be expressed as negative energy contributions, while task-importance penalties remain positive.

We impose a sparsity constraint on the number of pruned filters:
\begin{equation}
\sum_{i=1}^N p_i = K,
\end{equation}
where $K$ denotes the target number of \emph{pruned} filters; equivalently, the number of retained filters is $N-K$.  
Rather than enforcing this constraint via a hard quadratic penalty term, we satisfy it dynamically through a binary search on the capacity incentive coefficient during the optimization phase, as detailed in Section \ref{subsec:hyperparameters_and_sparsity}.

Our objective is to identify a binary pruning mask $\mathbf{p}$ such that the retained subset preserves task performance while satisfying the sparsity constraint.  
We first outline the baseline heuristic, then introduce progressively refined QUBO formulations that capture task sensitivity and pairwise score interactions, and finally describe the two-stage optimization pipeline using a Tensor-Train (TT) optimizer.

\subsection{Classic QUBO (L1-QUBO)}
We formulate pruning as a combinatorial optimization problem using a Quadratic Unconstrained Binary Optimization (QUBO):
\begin{equation}
\min_{\mathbf{p}\in\{0,1\}^N} \; \mathbf{p}^\top \mathbf{Q}\,\mathbf{p},
\label{eq:qubo_general}
\end{equation}
where $\mathbf{Q}$ encodes both linear importance and quadratic score interactions among filters.  

Following the foundational L1-based formulation introduced by Wang et al. \cite{wang2026quantum}, the per-filter score is computed as the mean absolute weight per filter:
\begin{equation}
I_i^{\ell_1} = \frac{1}{C_{\text{in}}^{(\ell)} \cdot k^{(\ell) 2}} \sum |W_i|,
\end{equation}
where $C_{\text{in}}^{(\ell)}$ is the number of input channels for the layer $\ell$ to which filter $i$ belongs, and $k^{(\ell)}$ is the spatial dimension of the square kernel. The L1-derived pairwise interaction between filters is then approximated via the outer product of these scores \cite{wang2026quantum}:
\begin{equation}
A_{ij} = I_i^{\ell_1} \, I_j^{\ell_1}.
\end{equation}

To incorporate the effect of model capacity into the objective, we adopt the per-filter capacity term $D_i$ proposed in \cite{wang2026quantum}, which reflects the relative parameter footprint of each filter. The number of parameters for filter $i$ in layer $\ell$ is given by:
\begin{equation}
N_i = C_{\text{in}}^{(\ell)} \cdot k^{(\ell) 2}.
\end{equation}
The total parameter-bit budget across the network is computed as:
\begin{equation}
S = \sum_{j=1}^{N} N_j \cdot b_{\max},
\end{equation}
where $b_{\max}$ denotes the maximum bit-width. The capacity term is then defined as the fraction of the total budget:
\begin{equation}
D_i = \frac{N_i \cdot b_{\max}}{S}.
\end{equation}
Since $b_{\max}$ is constant across all filters, this formulation effectively reduces to a normalized parameter count:
\begin{equation}
D_i = \frac{N_i}{\sum_{j=1}^{N} N_j}, \quad \text{such that} \quad \sum_i D_i = 1.
\end{equation}
Thus, $D_i$ represents the fraction of the total model capacity associated with filter $i$. 

The resulting QUBO matrix is constructed as:
\begin{align}
Q_{ii} &= A_{ii} - \gamma D_i, \\
Q_{ij} &= 2 A_{ij}, \quad i<j,
\end{align}
where $\gamma$ is a tunable hyperparameter. Because our objective is minimized and setting $p_i=1$ corresponds to pruning a filter, the negative term $-\gamma D_i$ lowers the overall energy. Therefore, this term acts as a \emph{capacity-driven pruning incentive}, actively rewarding the removal of larger filters rather than penalizing them.

This baseline L1-QUBO formulation is purely weight-based and task-agnostic, and generally underperforms compared to gradient-based heuristics such as Taylor pruning \cite{wang2026quantum}.

\subsection{Gradient-Aware QUBO}
To introduce task sensitivity into the QUBO objective, we inject the first-order Taylor importance
into the QUBO diagonal while keeping the pairwise score-interaction matrix unchanged. Concretely, let $I_i^{\text{Taylor}}$ denote the per-filter
Taylor importance. The QUBO entries are formed as:
\begin{align}
Q_{ii} &= \beta_{\mathrm{diag}}\,A_{ii} + \alpha\,I_i^{\text{Taylor}} - \gamma\,D_i, \\
Q_{ij} &= 2\,\beta_{\mathrm{off}}\,A_{ij}, \qquad i<j,
\end{align}
where $A$ encodes the pairwise score-interactions (e.g., weight-based correlations), $D_i$ represents the capacity incentive,
and $\alpha,\beta_{\mathrm{diag}},\beta_{\mathrm{off}},\gamma$ are tunable hyperparameters.
Note that the Taylor term enters the linear/diagonal part of the QUBO: the off-diagonals
remain proportional to $A_{ij}$. 

Under our minimization convention ($p_i=1$ means pruned), the positive term $+\alpha\,I_i^{\text{Taylor}}$ mathematically penalizes the removal of highly important filters. This design lets the solver balance task sensitivity (via the Taylor penalty) against pairwise score-interactions (via $A$) and capacity constraints, enabling globally coordinated pruning decisions beyond the greedy Taylor ranking \cite{molchanov2019importance,wang2026quantum}.

\subsection{Hybrid QUBO Framework and Activation Similarity}

Parameter redundancy captured by $A_{ij}$ models pairwise interactions between filter-importance scores, but it does not directly characterize functional redundancy in feature representations. Two filters can exhibit low parameter interaction while producing highly similar activations. To explicitly account for this functional redundancy, the general Hybrid QUBO framework is capable of incorporating activation-based similarity \cite{he2019filter,singh2020leveraging,geng2022pruning,wang2021convolutional,shaikh2025dynamic}.

For filters $i$ and $j$ within the same layer, let $A_i$ and $A_j$ denote their mean activation responses across a representative input sample. We define the normalized activation correlation:

\begin{equation}
S_{ij} \;=\; 
\frac{\langle A_i, A_j\rangle}
{\|A_i\|_2 \, \|A_j\|_2}
\;\in [-1,1].
\end{equation}

Positive values of $S_{ij}$ indicate functional redundancy, while negative values indicate diversity. To discourage the simultaneous selection of highly redundant filters, we introduce an additive quadratic penalty proportional to their similarity. The off-diagonal QUBO entries become:

\begin{equation}
Q_{ij} \;=\; 
2\,\beta_{\text{off}}\,A_{ij}
\;+\;
\lambda \,\max(0, S_{ij}),
\quad i<j,
\label{eq:qubo_actsim}
\end{equation}

where $\lambda \ge 0$ controls the strength of the activation redundancy penalty. Only positive correlations contribute to the penalty term, ensuring that diverse or negatively correlated filters are not artificially rewarded.

Under this formulation, if two filters exhibit high activation similarity ($S_{ij} \approx 1$), selecting both increases the quadratic cost, encouraging the solver to retain at most one of them. In contrast, dissimilar filters incur no additional penalty beyond the score-based interaction $A_{ij}$.

The general Hybrid QUBO framework is thus designed to balance task-aware sensitivity with activation redundancy. As detailed in the experimental sections, for the reported principal configuration evaluated in this paper, validation-based coefficient selection retained the Taylor and Weight-Fisher sensitivity terms. The functional activation-similarity coefficient $\lambda$ was optimized to a small non-zero value, contributing a fractional but measurable refinement to the final combinatorial objective.

\subsection{Weight-Fisher Sensitivity}
\label{subsec:fisher_methods}

The Taylor term used above is a first-order sensitivity proxy based on the first-order approximation of the loss change associated with perturbing the filter parameters. This is useful, but it does not fully capture how a filter behaves across the data distribution. In particular, near a well-trained solution, signed first-order gradients may partially cancel when averaged over different inputs, even if a filter is consistently sensitive on individual examples. Formally, for a parameter variable $\theta$,
\begin{equation}
\mathbb{E}_{(x,y)}\!\left[\frac{\partial \mathcal{L}(x,y)}{\partial \theta}\right] \approx 0,
\quad
\mathbb{E}_{(x,y)}\!\left[\left(\frac{\partial \mathcal{L}(x,y)}{\partial \theta}\right)^2\right] \not\approx 0.
\end{equation}
This motivates augmenting the QUBO diagonal with a second-moment term that measures not only average directional effect, but also how strongly the loss fluctuates with respect to a filter over different inputs \cite{lecun1989optimal,hassibi1992second,liu2021group,navarretescalable}.

\paragraph{Why Fisher?}
The quantity
\[
F_{\theta} \;=\; \mathbb{E}\!\left[\left(\frac{\partial \mathcal{L}}{\partial \theta}\right)^2\right]
\]
is the diagonal of the empirical Fisher information associated with the loss. Strictly speaking, for a general task loss this is an empirical Fisher / squared-gradient proxy rather than the exact Fisher information of a probabilistic model.
It can be interpreted as a data-distribution-aware measure of how sensitive the objective is to perturbations of $\theta$. While the Taylor pruning scores a filter using a first-order approximation of loss change at the current point, the Fisher term captures the second moment of the gradient signal across examples and therefore provides complementary information. In practice, this is especially useful when first-order signals are weak or unstable after averaging.

\paragraph{Weight-Fisher.}
For a scalar weight $w_k$, we define the diagonal empirical Fisher proxy
\[
F_k \;=\; \mathbb{E}\!\left[\left(\frac{\partial \mathcal{L}}{\partial w_k}\right)^2\right].
\]
For a convolutional filter $i$ with parameter set $\mathcal{P}_i$, we aggregate these sensitivities as
\begin{equation}
I_i^{\mathrm{Fisher}}
\;=\;
\sum_{k \in \mathcal{P}_i} w_k^2 \, F_k .
\label{eq:weight_fisher}
\end{equation}
This score is a second-moment sensitivity proxy associated with the potential effect of perturbing/removing the filter: large-magnitude weights that also exhibit high squared-gradient sensitivity receive larger importance.

\paragraph{Practical estimation.}
Fisher sensitivity is estimated from a calibration stream of mini-batches using one backward pass per batch.
We compute the per-filter quantity in Eq.~\eqref{eq:weight_fisher} from the current convolution weights and gradients, and accumulate it with an exponential moving average (EMA) across batches.

\paragraph{Integration into the QUBO.}
After normalization, the Fisher term is injected into the QUBO diagonal either as a replacement for, or in combination with, the Taylor term:
\begin{equation}
Q_{ii}
=
\beta_{\mathrm{diag}} A_{ii}
+
\alpha_T I_i^{\mathrm{Taylor}}
+
\alpha_F I_i^{\mathrm{Fisher}}
-
\gamma D_i.
\label{eq:fisher_qubo_diag}
\end{equation}
The off-diagonal terms remain unchanged relative to the chosen QUBO variant. In this way, Taylor contributes a first-order local signal, while Fisher contributes a distribution-aware second-moment signal, and the combinatorial formulation balances both against score interactions and compression incentives.

\subsection{Normalization and energy scaling}
\label{subsec:normalization}

The raw QUBO components arising from score interactions ($A$), importance ($I$), and capacity incentives ($D$)
can differ by several orders of magnitude, leading to ill-conditioned energy landscapes and unstable optimization.
To ensure balanced contributions and solver stability, we apply component-wise normalization prior to QUBO construction.

\paragraph{Importance normalization.}
When filter parameter counts $N_i$ are available, Taylor importance scores are first normalized as
\begin{equation}
\tilde I_i = \frac{I_i^{\mathrm{Taylor}}}{N_i},
\end{equation}
which prevents filters with larger receptive fields from being unfairly penalized due to scale alone.

\paragraph{Variance-based scaling.}
Each QUBO component is then scaled independently to unit variance using the empirical standard deviation of its magnitude:
\begin{align}
\hat A_{ii} &= \frac{A_{ii}}{\mathrm{std}(|A_{ii}|) + \epsilon}, \\
\hat A_{ij} &= \frac{A_{ij}}{\mathrm{std}(|A_{ij}|_{A_{ij}\neq 0}) + \epsilon}, \quad i \neq j, \\
\hat I_i &= \frac{\tilde I_i}{\mathrm{std}(|\tilde I|) + \epsilon}, \\
\hat D_i &= \frac{D_i}{\mathrm{std}(|D|) + \epsilon},
\end{align}
where $\epsilon$ is a small constant for numerical stability.
Diagonal and off-diagonal elements of $A$ are scaled separately to account for their distinct statistical roles.

\paragraph{Spectral norm capping.}
To further stabilize optimization, the implementation supports an optional spectral-norm cap on the normalized interaction matrix $\hat A$ to enforce
\begin{equation}
\|\hat A\|_2 \le 1,
\end{equation}
using a power-iteration estimate of the spectral norm.
In the reported experiments, this spectral norm capping was applied to prevent quadratic terms from dominating the energy and to improve robustness for the simulated annealing solvers.

After normalization, the hyperparameters $\alpha_T$, $\alpha_F$, $\beta_{\mathrm{diag}}$, $\beta_{\mathrm{off}}$, and $\gamma$
control relative trade-offs rather than compensating for raw scale differences, improving interpretability of hyperparameters and empirical transferability across models.

\subsection{Hyperparameter Optimization and Dynamic Capacity Search}
\label{subsec:hyperparameters_and_sparsity}

\paragraph{Handling the Sparsity Constraint ($K$).}
A common way to encode hard sparsity constraints in QUBO is to introduce a quadratic cardinality penalty term, such as $\eta (\sum_i p_i - K)^2$, directly into the QUBO matrix. Such penalties introduce additional quadratic terms and can substantially alter the energy landscape; in our experiment, the hard-penalty formulation also resulted in a larger QUBO and longer measured search time. Instead, we leverage the capacity incentive scalar ($\gamma$) to dynamically control the pruning percentage. Since $\gamma$ scales the capacity reward $D_i$, it effectively acts as a tunable threshold for sparsity: higher values of $\gamma$ uniformly increase the incentive to prune. To achieve the target sparsity without introducing an explicit quadratic cardinality penalty, we use a binary search over $\gamma$ to identify a capacity incentive that empirically yields the target cardinality. In each iteration of the binary search, we solve the unconstrained QUBO; we adjust $\gamma$ up or down until the solver returns a pruning mask that matches our target cardinality $K$. If multiple $\gamma$ values yield the exact target $K$, the framework selects the configuration with the lowest surrogate QUBO energy.

\paragraph{Hyperparameter Search Landscape.}
Our comprehensive objective function relies on balancing several competing terms: linear importance ($\alpha_T$ and/or $\alpha_F$), diagonal redundancy ($\beta_{\mathrm{diag}}$), off-diagonal score-interaction ($\beta_{\mathrm{off}}$), and activation similarity ($\lambda$). To find the optimal energy landscape for the solver, we perform a random search over predefined coefficient ranges. For each sampled configuration, we utilize the aforementioned binary search on $\gamma$ to ensure the resulting mask meets the exact sparsity constraint, and then evaluate the proxy configuration's performance.

\begin{figure}[t!]
\centering
\includegraphics[width=\columnwidth]{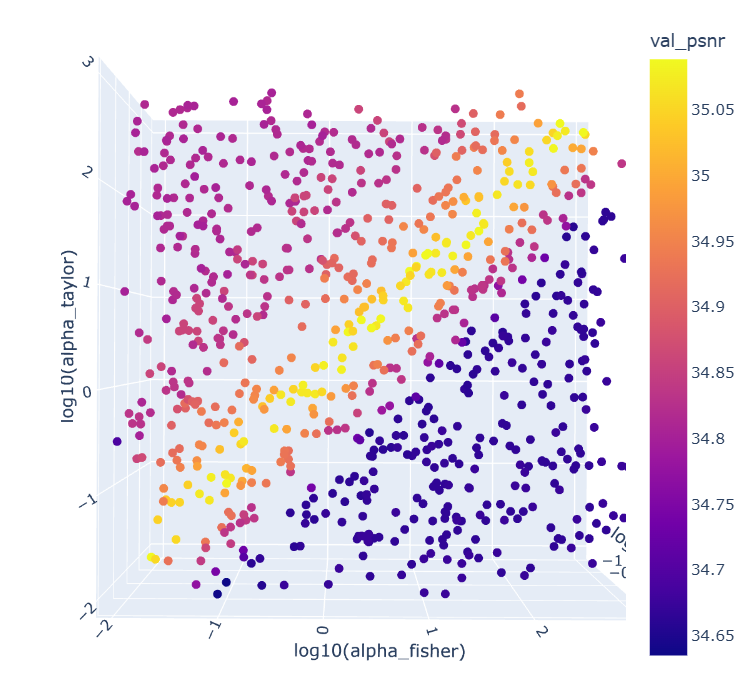}
\caption{\textbf{Hyperparameter search landscape showing validation PSNR across sampled coefficient configurations. Each point represents a sampled configuration evaluated at a fixed sparsity. High-performing configurations appear in broad regions of the sampled coefficient space, illustrating sensitivity to the relative balance between objective terms.}}
\label{fig:hyper_landscape}
\end{figure}

As illustrated in Figure \ref{fig:hyper_landscape}, our empirical hyperparameter search over multiple 3D slices reveals that high-performing configurations appear in broad regions of the sampled coefficient space. This demonstrates that the success of the QUBO formulation is sensitive to the relative balance between the competing energy terms. Identifying these broad high-performing regions motivates component-wise normalization (to keep the search space well-conditioned) and systematic hyperparameter tuning prior to final mask selection.

\subsection{Two-stage QUBO $\rightarrow$ Tensor-Train Refinement}
Even with a task-aware formulation, the QUBO objective provides a quadratic surrogate for the downstream pruning quality. The landscape of the true evaluation metric (e.g., PSNR) is accessed through function evaluations and is non-differentiable with respect to the discrete binary pruning mask $\mathbf{p}$. To reduce the gap between the QUBO proxy objective and the downstream evaluation metric, we employ a two-stage optimization strategy:

\begin{enumerate}
    \item \textbf{Stage 1 — QUBO optimization:}  
    We solve the chosen QUBO formulation to obtain an initial, high-quality pruning mask $\mathbf{p}_{\text{QUBO}}$. This stage searches the discrete solution space using the proposed QUBO proxy objective.
    
    \item \textbf{Stage 2 — Tensor-Train (TT) refinement:}  
    We use $\mathbf{p}_{\text{QUBO}}$ to initialize a gradient-free probabilistic optimizer based on the PROTES (PRobabilistic Optimization with TEnsor Sampling) framework \cite{batsheva2023protes,oseledets2011tensor,cichocki2017tensor}. 
    
    PROTES is designed for high-dimensional discrete spaces. Optimizing directly over the $2^N$ possible binary masks is computationally intractable. Instead of operating directly on the discrete variables, the TT optimizer maintains and optimizes a parameterized probability distribution over the entire search space using a Tensor Train (TT) format. For a binary mask $\mathbf{p} \in \{0, 1\}^N$, the probability of a configuration is represented by a tensor-train contraction of the corresponding slices of the TT cores. The optimization proceeds iteratively:
    \begin{itemize}
        \item \textbf{Sampling and Exploration:} The algorithm samples a batch of candidate masks from the current TT distribution. To prevent premature collapse to local optima, we explicitly inject the $\mathbf{p}_{\text{QUBO}}$ seed into early sampling iterations, alongside $\epsilon$-uniform exploration mass and local mutations.
        \item \textbf{Evaluation:} Because the TT optimizer explores freely, it must be constrained to maintain the target compression ratio. We evaluate each candidate $\mathbf{p}$ using a penalized derivative-free objective:
        \begin{equation}
            f(\mathbf{p}) = \text{Metric}(\mathbf{p}) - \mu_K \left| \sum_{j=1}^N p_j - K \right|,
        \end{equation}
        where $\text{Metric}$ is the true downstream evaluation (e.g., PSNR on a validation subset) and $\mu_K$ is the cardinality penalty coefficient that heavily discourages configurations deviating from the target sparsity $K$. By setting $\mu_K$ sufficiently high, the PROTES optimizer quickly learns to sample only valid configurations, effectively acting as a soft projection mechanism to ensure exact-$K$ feasibility.
        \item \textbf{Update:} The optimizer identifies the top-performing candidates (``elites'') from the batch. The tensor cores are then updated via gradient ascent (e.g., using the Adam optimizer) to maximize the log-likelihood of generating these elite configurations.
    \end{itemize}
\end{enumerate}

In summary, these results indicate that QUBO initialization can provide an effective starting point for subsequent derivative-free refinement. The QUBO stage provides a computationally efficient task-aware initialization, while PROTES subsequently searches for masks that improve the downstream metric \cite{batsheva2023protes,oseledets2011tensor,cichocki2017tensor}.

% --- Experimental Setup ---
\section{Experimental Setup}
\label{sec:setup}

This section defines the experimental environment used to evaluate the proposed framework. We describe the dataset and architectures, establish the pruning baselines and proposed variants, and specify the controlled experimental protocol, random seeds, solver settings, and Tensor-Train refinement configuration.

\subsection{Dataset and Model}
We conduct our primary experiments on the task of image denoising using the \textbf{SIDD Small sRGB} subset of the Smartphone Image Denoising Dataset \cite{abdelhamed2018high}, a high-quality dataset of real-world noisy images from smartphone cameras. Our model is a \textbf{Half-UNet} architecture with 2,368 prunable output filters, a common and effective model for this task, standing alongside other standard restoration architectures \cite{ronneberger2015u,zhang2017beyond,zamir2022restormer,lu2022half,chen2022simple}. All denoising models are evaluated using Peak Signal-to-Noise Ratio (PSNR) and Structural Similarity Index Measure (SSIM).

To assess cross-task transferability, we additionally evaluate the formulation on an image classification task using a CIFAR-style \textbf{ResNet-20} trained on the \textbf{CIFAR-10} dataset. For the classification model, top-1 accuracy is used as the primary metric.

\subsection{Baselines and Proposed Methods}
To evaluate the effectiveness of our approach, we compare the following pruning methods:
\begin{itemize}
    \item \textbf{Baseline (Unpruned):} The original, dense network model.
    \item \textbf{Classic L1-QUBO:} The QUBO formulation from \cite{wang2026quantum} using the task-agnostic L1-norm of the weights to construct the score-interaction and importance components.
    \item \textbf{Taylor (Greedy):} The strong standard baseline \cite{molchanov2019importance}, which greedily removes the least important filters based on first-order Taylor expansion approximations.
    \item \textbf{Hybrid QUBO (Ours):} Our general combinatorial formulation. For the principal configuration, validation-based selection retained Taylor and Fisher sensitivity, and optimized activation-similarity to a small fractional value.
    \item \textbf{TT Refinement (Ours):} Our two-stage refinement strategy, where the Tensor-Train (TT) optimizer performs gradient-free probabilistic optimization to search for improved pruning masks.
\end{itemize}

\subsection{Experimental Design}
To ensure a controlled comparison, we distinguish the dataset split from the stochasticity of the optimization procedure. The train/validation/test partition is fixed throughout the statistical experiments using a dataset split seed of 42. The optimization seed is then varied independently across repeated runs.

For the full-model statistical evaluation, we use the optimization seeds \{101, 202, 303, 404, 505, 606, 707, 808, 909, 1010\}. All methods within a given run use the same fixed dataset partition and are evaluated on the same held-out test set. This protocol isolates variation caused by the stochastic optimization procedure from variation caused by data partitioning.

For sensitivity estimation in the revised statistical protocol, the calibration data are processed deterministically, without stochastic training-time augmentation or data-loader shuffling. The resulting importance estimates are therefore evaluated under a fixed input ordering and fixed model parameters. For each optimization seed, the corresponding pruning mask is then applied to a freshly loaded copy of the dense model before test evaluation.

For the primary full-model Half-UNet experiment, the target pruning cardinality is $K=858$ filters out of 2,368 prunable filters, corresponding to $\frac{858}{2368} \times 100 \approx 36.23\%$ filter pruning.

To systematically analyze the specific behavior of the TT Refinement stage under varying search space sizes, we also conduct \textbf{controlled sub-problem experiments}. We isolate 4, 8, and 16 representative prunable layers, containing $N=256$, $512$, and $1024$ binary filter variables, respectively, and additionally evaluate the full model with $N=2368$. For the Tensor-Train scaling experiments reported in Table~\ref{tab:tt_scaling}, the pruning cardinality is $K=\operatorname{round}((858/2368)N)$, giving $K=93$, $186$, $371$, and $858$, respectively. The QUBO seed and its refined masks use the same target cardinality within each configuration.

\subsection{Reproducibility and Statistical Protocol}
The revised statistical analysis separates the sources of randomness: the dataset split seed, the optimization seed, the QUBO solver seed, and the Tensor-Train optimizer seed. The dataset partition is fixed (seed 42), whereas the optimization seed is varied across independent runs. For the QUBO experiments, the simulated-annealing solver seed was set equal to the optimization seed for reproducible repeated optimization. PROTES uses specific seeds (9999, 10000, 10001) for the scaling runs. The test set was strictly excluded from coefficient selection or model/mask selection.

We report the mean and standard deviation of PSNR and SSIM across the repeated optimization runs. For paired comparisons, we compute 95\% confidence intervals for the paired differences. A paired $t$-test is used when the paired differences satisfy the normality criterion; otherwise, the Wilcoxon signed-rank test is used. Paired Cohen's $d$ is additionally reported as an effect-size measure. When multiple comparisons are performed, $p$-values are adjusted using the Holm procedure. The values reported in our statistical tables are unadjusted (raw) $p$-values.

% --- Results ---
\section{Results}
\label{sec:results}

This section evaluates the proposed framework from complementary perspectives. We first report full-model pruning performance and the corresponding component ablation. We then quantify statistical robustness across optimization seeds and validate the Dynamic Capacity Search mechanism. Next, we examine the scaling behavior of Tensor-Train refinement and evaluate cross-task transferability on the CIFAR-10 classification task. Finally, we report optimization complexity and current masked-model deployment measurements.

\subsection{Main Full-Model Comparison and Ablation}
We evaluate pruning over the complete set of 2,368 prunable filters in the Half-UNet architecture. The target is to remove $K=858$ filters, corresponding to 36.23\% filter sparsity.

To isolate the contributions of individual components in the proposed framework, we conducted a controlled ablation study at this target sparsity. As detailed in Table \ref{tab:ablation}, the dense model achieves a PSNR of 37.2633~dB and an SSIM of 0.931648 on the fixed test split. Relying solely on score-interaction and capacity incentives (Classic L1-QUBO) yields poor task performance (24.6539~dB). Integrating first-order Taylor sensitivity (QUBO + Taylor) recovers significant performance (34.7891~dB), performing comparably to the greedy Taylor baseline (34.8080~dB). 

Incorporating the empirical Fisher sensitivity proxy (QUBO + Taylor + Fisher) further improves the resulting pruning configuration, surpassing the greedy baseline to achieve 35.0652~dB. The final Full Hybrid QUBO configuration incorporates a small, empirically optimized functional activation similarity term ($\lambda$) that further improves the resulting pruning configuration to 35.0717~dB. These ablation results indicate that task-aware sensitivity terms substantially affect the quality of the resulting pruning masks, while the relative benefit of the individual components depends on the selected objective configuration.

\begin{table*}[t]
\caption{\textbf{Ablation of Hybrid QUBO Components at 36.23\% filter sparsity. Values are mean $\pm$ standard deviation.}}
\label{tab:ablation}
\setlength{\tabcolsep}{6pt}
\begin{center}
\begin{tabular}{lrrrr}
\toprule
\textbf{Method} & \textbf{PSNR (dB)} & $\mathbf{\Delta}$ PSNR vs Taylor & \textbf{SSIM} & $\mathbf{\Delta}$ SSIM vs Taylor \\
\midrule
Dense Baseline & 37.2633 & $--$ & 0.931648 & $--$ \\
Taylor (Greedy) & $34.8080 \pm 0.0000$ & Ref & $0.896398 \pm 0.000000$ & Ref \\
Classic L1-QUBO & $24.6539 \pm 0.1223$ & $-10.1541$ & $0.624152 \pm 0.010379$ & $-0.272246$ \\
QUBO + Taylor & $34.7891 \pm 0.0000$ & $-0.0189$ & $0.895821 \pm 0.000000$ & $-0.000577$ \\
QUBO + Taylor + Fisher & $35.0652 \pm 0.0000$ & $+0.2572$ & $0.912648 \pm 0.000000$ & $+0.016250$ \\
\textbf{Full Hybrid QUBO} & $\mathbf{35.0717 \pm 0.0000}$ & $\mathbf{+0.2637}$ & $\mathbf{0.913002 \pm 0.000000}$ & $\mathbf{+0.016604}$ \\
\bottomrule
\end{tabular}
\end{center}
\end{table*}

\subsection{Statistical Robustness}
To assess the robustness of the pruning methods to stochastic optimization, we performed ten independent runs using different optimization seeds while keeping the dataset split fixed.

The proposed Hybrid QUBO consistently produced the same test-set performance across all ten seeds, with PSNR $=35.0717 \pm 0.0000$~dB and SSIM $=0.913002 \pm 0.000000$. Taylor pruning likewise produced an identical mask and corresponding performance under the deterministic calibration protocol. Classic L1-QUBO exhibited measurable variation across seeds, with PSNR $=24.6539 \pm 0.1223$~dB and SSIM $=0.624152 \pm 0.010379$. Thus, in this experiment, the stochasticity of the solver has a much larger effect on the Classic L1-QUBO configuration than on the task-aware Hybrid QUBO configuration.

Because the seed-level Taylor and Hybrid results are deterministic under the fixed protocol, we additionally evaluate their difference using paired per-image analysis on the fixed held-out test set for a representative run. Table~\ref{tab:ablation} reports PSNR computed from the mean squared error pooled within each 16-image test batch and then averaged across the four batches. Table~\ref{tab:per_image_statistics} instead computes PSNR separately for each of the 64 test images and analyzes the paired differences for optimization seed 101. Because the logarithm is nonlinear, the mean per-image PSNR difference need not equal the difference between the batch-aggregated PSNR values in Table~\ref{tab:ablation}.

\begin{table*}[t]
\caption{\textbf{Paired per-image statistical comparison on the fixed test set.}}
\label{tab:per_image_statistics}
\setlength{\tabcolsep}{6pt}
\begin{center}
\begin{tabular}{lrrrr}
\toprule
\textbf{Metric} & \textbf{Comparison} & \textbf{Mean Diff.} & \textbf{95\% CI} & \textbf{Raw $p$-value} \\
\midrule
PSNR & Hybrid vs.\ Taylor & $+0.3068$ & $[0.2538,0.3598]$ & $3.21\times10^{-17}$ \\
SSIM & Hybrid vs.\ Taylor & $+0.0166$ & $[0.0142,0.0190]$ & $3.53\times10^{-12}$ \\
\bottomrule
\end{tabular}
\end{center}
\end{table*}

The corresponding paired effect sizes are $d=1.45$ for PSNR and $d=1.74$ for SSIM when comparing Hybrid QUBO with Taylor. These results indicate that the performance differences are not only numerically observable but also large relative to the paired per-image variability.

\subsection{Dynamic Capacity Search}
To control the pruning cardinality without introducing an explicit quadratic cardinality penalty, we evaluate the Dynamic Capacity Search against a standard hard quadratic penalty formulation. 

\begin{table*}[t]
\caption{\textbf{Comparison of Dynamic Capacity Search vs. Hard Quadratic Penalty.}}
\label{tab:dcs_validation}
\setlength{\tabcolsep}{6pt}
\begin{center}
\begin{tabular}{lrrrrrr}
\toprule
\textbf{Approach} & \textbf{Mean Sparsity Err.} & \textbf{Max Err.} & \textbf{PSNR (dB)} & \textbf{SSIM} & \textbf{Time (s)} & \textbf{QUBO Terms} \\
\midrule
Dynamic Capacity Search & 0.0 & 0 & 36.0506 & 0.9178 & 31.82 & 33,280 \\
Hard Quadratic Penalty & 0.8 & 1 & 36.1230 & 0.9187 & 72.21 & 524,800 \\
\bottomrule
\end{tabular}
\end{center}
\end{table*}

As detailed in Table \ref{tab:dcs_validation}, in these experiments, Dynamic Capacity Search achieved exact cardinality control with zero maximum sparsity error, requires approximately 31.82 seconds, and generates only 33,280 QUBO terms. In contrast, the hard quadratic penalty yields up to a 1-filter error, requires 72.21 seconds, and generates 524,800 QUBO terms. This demonstrates that dynamic capacity search provides exact cardinality control while substantially reducing the number of QUBO terms and measured search time, at a small quality trade-off relative to the hard-penalty formulation.

\subsection{Tensor-Train Refinement Scaling}
To evaluate the scaling behavior and practical value of the Tensor-Train (PROTES) refinement stage, we analyzed its performance across sub-problems of increasing dimensionality as well as the full network. For each configuration, refinement was evaluated over 3 independent optimization runs using different random seeds (seeds 9999, 10000, and 10001) under a fixed evaluation budget of $m=20{,}000$.

\begin{table*}[t]
\caption{\textbf{Scale-dependent Tensor-Train refinement on Half-UNet. Values represent the mean and standard deviation over 3 independent refinement runs. The QUBO seed is fixed within each controlled configuration.}}
\label{tab:tt_scaling}
\setlength{\tabcolsep}{6pt}
\begin{center}
\begin{tabular}{lrrrcccc}
\toprule
\textbf{Problem} & \textbf{Variables ($N$)} & \textbf{Target $K$} & \textbf{QUBO Seed (dB)} & \textbf{TT Refined (dB)} & \textbf{Test Gain (dB)} & \textbf{Unique Evals} & \textbf{Runtime (s)} \\
\midrule
4 layers   & 256  & 93  & 37.0133 & $37.1122 \pm 0.0225$ & $+0.0989 \pm 0.0225$ & $\sim$19,913 & $\sim$2265 \\
8 layers   & 512  & 186 & 36.7477 & $36.9041 \pm 0.0334$ & $+0.1563 \pm 0.0334$ & $\sim$19,922 & $\sim$2277 \\
16 layers  & 1024 & 371 & 36.0506 & $36.5293 \pm 0.0368$ & $+0.4787 \pm 0.0368$ & $\sim$19,924 & $\sim$2574 \\
Full model & 2368 & 858 & 34.8571 & $35.6601 \pm 0.0081$ & $+0.8030 \pm 0.0081$ & $\sim$19,961 & $\sim$3023 \\
\bottomrule
\end{tabular}
\end{center}
\end{table*}

As reported in Table \ref{tab:tt_scaling}, PROTES improved upon the QUBO seed in all 12 evaluated runs. The observed magnitude of recovery increased with the dimensionality of the controlled search problems:
\begin{itemize}
    \item \textbf{Small sub-problems ($N \le 512$):} The QUBO proxy objective identifies robust discrete solutions, leaving the observed room for further improvement under the fixed PROTES budget relatively small ($+0.0989 \pm 0.0225$ dB for 4 layers and $+0.1563 \pm 0.0334$ dB for 8 layers).
    \item \textbf{Intermediate scale ($N = 1024$):} Seeding PROTES with the QUBO solution yields a noticeable improvement of $+0.4787 \pm 0.0368$ dB on the test set.
    \item \textbf{Full model ($N = 2368$):} In the largest discrete search space ($2^{2368}$), where the observed gap between the QUBO solution and the downstream metric is largest, the exact-$K$ PROTES refinement delivers its largest recovery, improving test PSNR by $+0.8030 \pm 0.0081$ dB over the QUBO seed.
\end{itemize}

\textit{Note: The TT scaling experiment uses a separate controlled evaluation protocol (including a smaller specific 32-image validation subset) and therefore its full-model QUBO seed test PSNR (34.8571 dB) is not identical to the seed used in the principal ten-seed statistical experiment (35.0717 dB). These results are not combined across protocols.}

\subsection{Cross-Task Evaluation on CIFAR-10}
To evaluate the cross-task transferability of the proposed optimization strategy, we applied the framework to an image classification task using a CIFAR-style ResNet-20 trained on the CIFAR-10 dataset. The dense checkpoint achieved a top-1 test accuracy of 92.190\%. The controlled pruning problem contained $N=672$ output-filter variables across 18 residual convolutions. 

We evaluated the methods at exact target nominal filter sparsities of 30\%, 50\%, and 70\%. The exact-K PROTES refinement \cite{batsheva2023protes} initialized by the Hybrid QUBO was compared against the greedy Taylor pruning baseline \cite{molchanov2019importance}. To comprehensively assess mask quality, we evaluated both the zero-shot performance (raw capacity of the mask without post-pruning fine-tuning or BatchNorm recalibration) and the post-fine-tuning convergence (after 30 epochs of standard retraining).

\begin{table*}[t]
\caption{\textbf{ResNet-20 on CIFAR-10 Test Accuracy and Weight Reduction at Exact Target Filter Sparsities.}}
\label{tab:resnet_generalization}
\setlength{\tabcolsep}{6pt}
\begin{center}
\begin{tabular}{llccc}
\toprule
\textbf{Filter Sparsity} & \textbf{Method} & \textbf{Zero-Shot Acc.} & \textbf{Fine-Tuned Acc.} & \textbf{Conv. Weight Reduction} \\
\midrule
\multirow{2}{*}{30\% ($K=202$)} & Taylor Baseline \cite{molchanov2019importance} & 17.43\% & 91.29\% & 34.05\% \\
                                & \textbf{Hybrid QUBO + PROTES} & \textbf{68.90\%} & \textbf{91.42\%} & \textbf{39.70\%} \\
\midrule
\multirow{2}{*}{50\% ($K=336$)} & Taylor Baseline \cite{molchanov2019importance} & 18.81\% & \textbf{89.67\%} & 54.88\% \\
                                & \textbf{Hybrid QUBO + PROTES} & \textbf{51.30\%} & 88.32\% & \textbf{64.58\%} \\
\midrule
\multirow{2}{*}{70\% ($K=470$)} & Taylor Baseline \cite{molchanov2019importance} & 12.26\% & 84.63\% & 73.85\% \\
                                & \textbf{Hybrid QUBO + PROTES} & \textbf{39.67\%} & \textbf{84.72\%} & \textbf{79.39\%} \\
\bottomrule
\end{tabular}
\end{center}
\end{table*}

As detailed in Table \ref{tab:resnet_generalization}, the combinatorial optimization strategy demonstrates distinct advantages in two regimes:

First, in the strict \textbf{zero-shot setting}, the Hybrid QUBO + PROTES framework preserves representation fidelity substantially better than the greedy baseline. At 30\% filter sparsity, the optimized mask retains 68.90\% accuracy compared to Taylor's 17.43\% (+51.47 percentage points). At 70\% sparsity, it retains 39.67\% versus Taylor's 12.26\% (+27.41 percentage points). This indicates that the task-aware interaction modeling selects configurations that are fundamentally less destructive to the un-retrained computational graph.

Second, in the \textbf{post-fine-tuning setting}, 30 epochs of gradient descent act as an equalizer, with both methods recovering highly competitive accuracies (e.g., $\sim$91.4\% at 30\% sparsity). However, because the Hybrid QUBO formulation actively incorporates a capacity-reduction incentive ($D_i$), it inherently prioritizes the removal of larger filters containing more parameters. Consequently, at identical nominal filter sparsities, the combinatorial approach yields a physically lighter model. For example, at 50\% filter sparsity, the QUBO-derived mask eliminates 64.58\% of the actual convolutional weight volume, compared to only 54.88\% for the Taylor mask. This $\sim$9.7 percentage point difference in actual parameter reduction contextualizes the minor accuracy variations observed after fine-tuning, demonstrating that the combinatorial approach yields superior accuracy-to-parameter trade-offs.

\subsection{Computational Cost}
The computational cost of the proposed framework is dominated by data-dependent importance estimation rather than QUBO matrix construction itself. For the full-model Hybrid QUBO experiment, Taylor and Fisher importance collection required approximately 207.01~s under the deterministic calibration protocol.

The construction of the QUBO tensor representation required only 0.06~s, while the dynamic capacity search required 10.85~s. Construction of the final QUBO dictionary required an additional 0.89~s, followed by 23.62~s for the final simulated-annealing solve. Thus, the major computational contribution of the proposed optimization pipeline is the gradient-based sensitivity estimation stage, while the matrix assembly and final combinatorial solve constitute smaller portions of the measured optimization time. Furthermore, computing the activation similarity $S_{ij}$ requires only a forward pass and a cosine similarity computation, adding negligible overhead compared to gradient backpropagation.

The quadratic structure of QUBO also introduces an important scalability consideration. For $N$ binary pruning variables, a dense quadratic formulation contains up to $O(N^2)$ pairwise interactions. Consequently, increasing the number of simultaneously optimized filters rapidly increases both memory requirements and the computational cost of constructing and solving the corresponding QUBO. This quadratic growth motivates the controlled sub-problem experiments and the use of TT-based refinement only where the resulting derivative-free search remains computationally manageable.

\subsection{Masked-Model Deployment Profiling}
Filter sparsity alone does not directly determine the actual deployment efficiency of a pruned neural network. Therefore, we distinguish between theoretical reductions implied by a pruning mask and the measured runtime of the current masked PyTorch implementation.

For the dense Half-UNet, we measure 68,899 parameters, approximately $2.617\times10^9$ convolutional FLOPs for a $256\times256$ input, and an inference latency of 8.99~ms on the evaluated GPU and 136.38~ms on CPU.

For the Hybrid QUBO mask, the current implementation retains the original tensor dimensions and applies binary masks to output filters. Consequently, the masked model is not structurally compacted. Its measured latency is 15.17~ms on GPU and 139.55~ms on CPU, corresponding to approximately $0.59\times$ and $0.98\times$ the dense-model speed, respectively.

These measurements demonstrate that applying a pruning mask alone does not guarantee hardware acceleration. Actual deployment gains require structural reconstruction of the affected layers and propagation of channel dependencies through the computational graph. Therefore, the present latency measurements are reported as \emph{masked-model profiling} rather than as evidence of structural deployment acceleration.

\begin{table*}[t]
\caption{\textbf{Current masked-model profiling for the full-model Hybrid QUBO configuration.}}
\label{tab:masked_profile}
\setlength{\tabcolsep}{6pt}
\begin{center}
\begin{tabular}{lcc}
\toprule
\textbf{Metric} & \textbf{Dense} & \textbf{Hybrid QUBO Masked} \\
\midrule
Parameters & 68,899 & -- \\
Convolutional FLOPs & $2.617\times10^9$ & -- \\
GPU latency (ms) & 8.99 & 15.17 \\
CPU latency (ms) & 136.38 & 139.55 \\
\bottomrule
\end{tabular}
\end{center}
\end{table*}
\noindent
The parameter and FLOP entries of the masked implementation are intentionally left unspecified in the main table until the structural dependency-aware accounting is fully validated. The binary pruning mask itself removes 858 of 2,368 prunable filters (36.23\%), but this percentage should not be interpreted as an equivalent parameter or FLOP reduction.

% --- Discussion ---
\section{Discussion}
\label{sec:discussion}

This section interprets the experimental findings in terms of task-aware combinatorial optimization, robustness to stochastic search, performance vs. computational cost, and practical deployment considerations. We distinguish experimentally demonstrated advantages from limitations and from claims requiring further validation.

The revised experiments provide evidence that the proposed task-aware QUBO formulation is more effective than the evaluated greedy Taylor and task-agnostic L1-QUBO baselines at the studied pruning target. At 36.23\% filter pruning, the Hybrid QUBO achieves 35.0717~dB PSNR and 0.913002 SSIM, compared with 34.8080~dB and 0.896398 for Taylor pruning.

\subsection{Task-Aware Combinatorial Optimization}
The difference between Classic L1-QUBO and the task-aware formulation is substantial. Classic L1-QUBO reaches only 24.6539~dB PSNR on average, whereas the Hybrid QUBO reaches 35.0717~dB. This result indicates that the choice of optimization objective is critical: the ability to solve a combinatorial problem globally does not by itself guarantee a useful pruning configuration. The objective must provide task-relevant information that correlates with the effect of pruning on denoising performance.

\subsection{Robustness to Optimization Seeds}
Under the fixed deterministic dataset protocol, the Hybrid QUBO produces identical test performance across the ten evaluated optimization seeds. This should be interpreted specifically as robustness to the stochasticity of the optimization procedure under fixed model parameters and fixed calibration data, rather than as evidence of robustness to changes in datasets, architectures, training procedures, or model initialization.

Taylor pruning also produces identical results because the importance estimation and ranking are deterministic under the fixed calibration stream. In contrast, Classic L1-QUBO shows measurable variation across optimization seeds, indicating greater sensitivity of its solution to the stochastic simulated-annealing search.

\subsection{Performance vs. Computational Cost}
An explicit trade-off emerges when comparing QUBO optimization to Tensor-Train refinement. QUBO is substantially cheaper—requiring approximately 242 seconds for the measured sensitivity estimation, capacity search, QUBO construction, and final simulated-annealing stages combined—but it yields a lower proxy-based solution. In contrast, TT refinement is much more computationally expensive, ranging from roughly 2265 seconds to over 3000 seconds for 20,000 evaluations. 

However, TT directly optimizes the downstream metric and increasingly improves the final solution as the problem dimension grows. This highlights a clear operational trade-off: QUBO provides a highly efficient task-aware initialization, while TT can be deployed when computational budgets permit extensive derivative-free refinement to reduce the gap between the QUBO proxy objective and the downstream evaluation metric.

\subsection{Scope and Practical Interpretation}
The current evidence supports the use of Hybrid QUBO as a task-aware pruning formulation for the evaluated Half-UNet denoising problem. The results do not yet establish universal superiority over all structured-pruning techniques or across different architectures and tasks. Likewise, deployment efficiency cannot be inferred from filter sparsity alone.

The Tensor-Train refinement stage should therefore be interpreted as an optional derivative-free optimization component whose practical value depends on the dimensionality of the controlled search problem. The completed controlled experiments characterize this scaling behavior, determining whether the additional optimization cost is justified at the largest search scale.

\subsection{Limitations}
Several limitations should be considered when interpreting the present results. First, the principal empirical evaluation remains centered on the SIDD image-denoising task and the Half-UNet architecture. Although we include an additional ResNet-20/CIFAR-10 evaluation, the experiments do not establish broad generalization across architectures and tasks.

Second, the current masked implementation does not physically remove channels from the model tensors. Therefore, theoretical sparsity and measured deployment latency should be treated separately until a structurally compact model is constructed and benchmarked.

Third, the proposed QUBO formulation has quadratic growth in the number of pairwise interactions. While restricting the interactions $A_{ij}$ and $S_{ij}$ strictly to intra-layer filter pairs yields a block-diagonal (or block-sparse) QUBO matrix that naturally manages this $O(N^2)$ growth, extending the formulation to capture arbitrary inter-layer coupling limits the practical size of a single global optimization problem and motivates decomposition or hierarchical strategies for larger networks.

Finally, we do not claim universal superiority over all state-of-the-art structured pruning methods (e.g., FPGM, Network Slimming, or CHITA). The current evaluation establishes that Hybrid QUBO outperforms the evaluated Taylor and L1-QUBO baselines, but a fair, controlled comparison across target sparsities with these advanced structural methods remains for future work.

% --- Conclusion ---
\section{Conclusion}
\label{sec:conclusion}
This work investigated structured neural network pruning as a combinatorial optimization problem and proposed a Hybrid QUBO formulation that combines task-aware sensitivity information with filter-interaction modeling. For the evaluated Half-UNet denoising model, the proposed method achieved higher PSNR and SSIM than the evaluated Taylor and L1-based QUBO baselines at the studied pruning target.

The revised multi-seed protocol further shows that the observed performance is reproducible under a fixed dataset partition and controlled optimization seeds. The computational analysis indicates that data-dependent importance estimation is the dominant component of the optimization cost, while QUBO construction and the final discrete optimization require comparatively smaller portions of the measured runtime.

The study also highlights an important distinction between pruning sparsity and deployment efficiency. Applying binary masks to dense tensors does not by itself guarantee structural parameter reduction or hardware acceleration. A deployment-oriented implementation therefore requires structural reconstruction and dependency-aware channel propagation.

Finally, the Tensor-Train refinement stage provides a mechanism to search for improved pruning masks using the downstream metric. Its practical value depends on the dimensionality of the discrete search space and the available evaluation budget. Broader evaluation across architectures, tasks, and structured-pruning baselines remains necessary to establish the generality of the proposed framework.

% --- Appendices ---
\appendices

\section{Reproducibility Details}

\subsection{Dataset and Preprocessing}
We use the SIDD Small sRGB subset of the Smartphone Image Denoising Dataset \cite{abdelhamed2018high}. Paired noisy and ground-truth images are processed into $256\times256$ patches.

For the revised statistical experiments, the partition into training, validation, and test sets is generated once using dataset split seed $42$ and then kept fixed for all optimization seeds. The resulting evaluation protocol contains 1,536 training samples, 64 validation samples, and 64 test samples in the loader configuration used for the reported experiments. No re-partitioning is performed when changing the optimization seed.

During training, random horizontal and vertical flips were applied with the implementation-specific augmentation probabilities described in the training code. Validation and test samples are evaluated without stochastic augmentation. For sensitivity estimation in the revised statistical protocol, we additionally use a deterministic calibration stream with fixed ordering and no stochastic augmentation.

\subsection{Random Seeds}
The experimental protocol distinguishes the dataset split seed from optimization-related random seeds. The dataset partition is fixed using seed $42$, while the principal repeated experiments use the optimization seeds
\[
\{101,202,303,404,505,606,707,808,909,1010\}.
\]
For each optimization run, the corresponding pruning procedure is initialized using the specified seed. The deterministic calibration stream remains unchanged across runs, allowing the effect of optimization stochasticity to be evaluated independently of changes in the calibration data.

For the QUBO experiments, the simulated-annealing solver seed was set equal to the optimization seed for reproducible repeated optimization. The Tensor-Train refinement stage uses specific initialization seeds (9999, 10000, 10001) for scaling evaluations when enabled. The test set was strictly excluded from coefficient selection or model/mask selection.

\subsection{CIFAR-10 ResNet-20 Training Details}
The CIFAR-style ResNet-20 model was trained on the CIFAR-10 dataset using a batch size of 128 for 160 epochs. We used SGD with a momentum of 0.9, weight decay of $5\times10^{-4}$, and an initial learning rate of 0.1, which was decayed by a factor of 0.1 at epochs 80 and 120. The evaluation batch size was set to 256. The Fisher calibration utilized 25 batches from the training set, and the PROTES validation subset comprised 512 samples. The specific coefficient configuration was selected independently for the CIFAR-10 experiment using its own validation protocol.

\subsection{Model Architecture}
Our denoising model is a Half-UNet with NAF (Nonlinear Activation Free) blocks \cite{lu2022half,chen2022simple} featuring 2,368 prunable output filters. Key components include:
\begin{itemize}
    \item Input convolution to project 3 channels to \texttt{FILTER} feature maps.
    \item Three encoder stages with two consecutive NAFBlocks each, followed by max-pooling.
    \item PixelShuffle-based upsampling with skip connections.
    \item NAFBlock features:
    \begin{itemize}
        \item Depthwise convolutions
        \item Simplified channel attention (SCA)
        \item SimpleGate activation
        \item LayerNorm2d normalization
        \item Learnable residual weights $\beta$ and $\gamma$
    \end{itemize}
    \item Final $1 \times 1$ convolution to produce 3-channel output
\end{itemize}

\subsection{Hyperparameter Optimization Protocol}
Hyperparameters including $\alpha_T$, $\alpha_F$, $\beta_{\mathrm{diag}}$, $\beta_{\mathrm{off}}$, and $\lambda$ were selected via random search. The search evaluated 800 configurations using log-uniform sampling over predefined ranges: $\alpha_T \in [0.01, 10.0]$, $\alpha_F \in [0.01, 10.0]$, $\beta_{\mathrm{diag}} \in [0.01, 10.0]$, $\beta_{\mathrm{off}} \in [10^{-4}, 0.2]$, and $\lambda \in [10^{-4}, 1.5]$. We evaluated the sampled configurations using a validation subset of 64 samples, with the highest validation PSNR used as the selection criterion. The test set was strictly excluded from this process; all coefficient and mask selections were performed exclusively using the designated validation subset. For the reported principal configuration, validation-based coefficient selection retained the Taylor and Fisher terms. The functional activation-similarity coefficient $\lambda$ was optimized to a small non-zero value, contributing a fractional but measurable refinement to the final combinatorial objective. The selected principal configuration used $\alpha_T=1.0$, $\alpha_F=0.5$, $\beta_{\mathrm{diag}}=1.0$, $\beta_{\mathrm{off}}=0.01$, and $\lambda=10^{-4}$.

\subsection{Training Procedure}
\begin{itemize}
    \item Optimizer: Adam
    \item Learning rate: $1 \times 10^{-3}$
    \item Scheduler: StepLR with step size 10 epochs, $\gamma=0.1$
    \item Batch size: 16
    \item Epochs: 100
    \item Device: GPU (CUDA)
\end{itemize}

The loss function is defined as:
\[
\mathcal{L}(x, y) = 100 - \text{PSNR}(x, y)
\]
where PSNR is computed as:
\[
\text{PSNR}(x, y) = 20 \log_{10} \frac{\text{data\_range}}{\sqrt{\text{MSE}(x, y)}}
\]

\subsection{Evaluation Metrics}
Performance was evaluated using:
\begin{itemize}
    \item Peak Signal-to-Noise Ratio (PSNR)
    \item Structural Similarity Index Measure (SSIM)
\end{itemize}

Average metrics for each epoch were recorded for both training and validation sets. The final model was saved and evaluated on the held-out test set.

\subsection{Training History}
\begin{figure}[t!]
\centering
\includegraphics[width=\columnwidth]{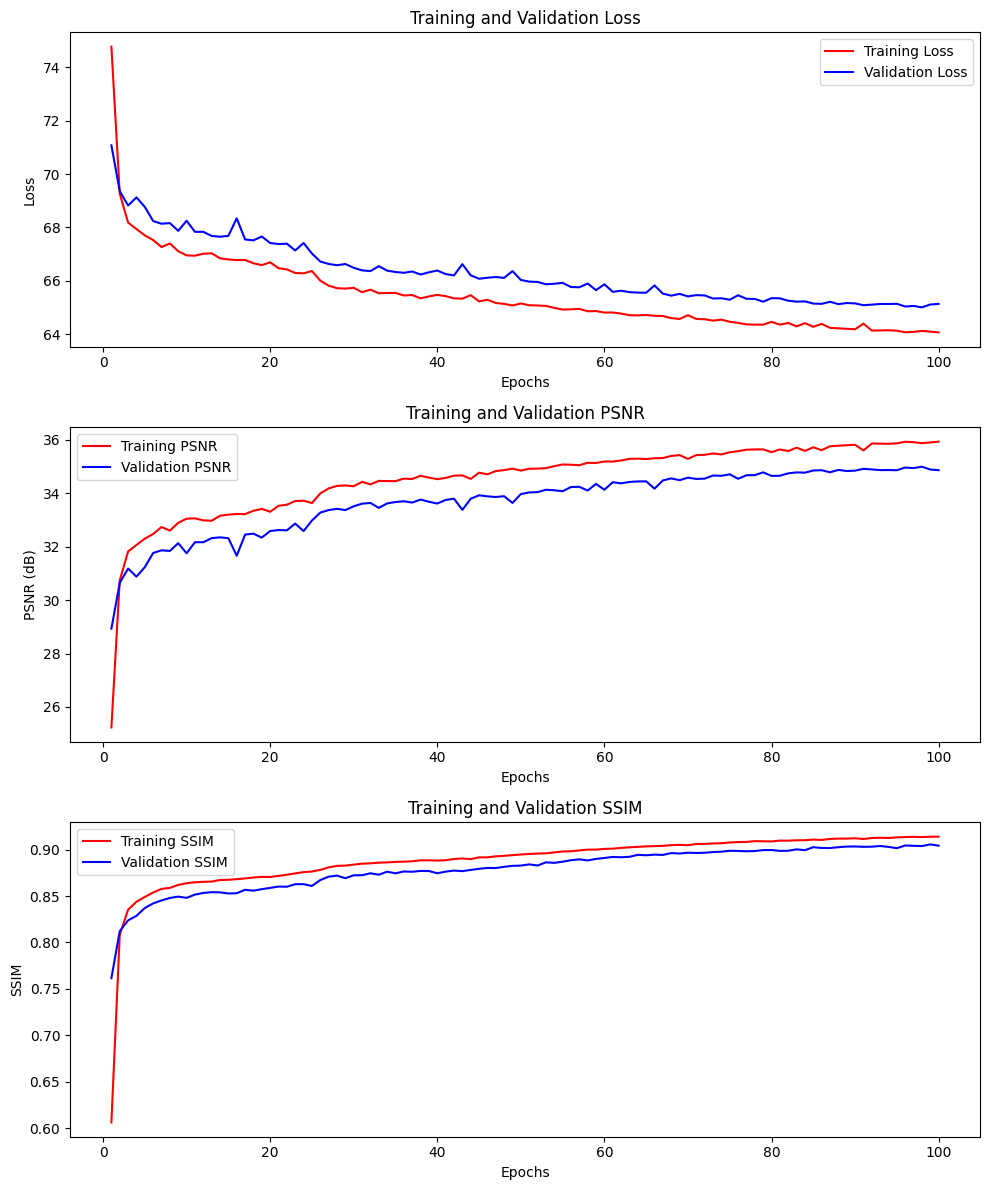}
\caption{\textbf{Training and validation history of the baseline Half-UNet model over 100 epochs. The plots display the stable convergence of the custom Loss (top), Peak Signal-to-Noise Ratio (middle), and Structural Similarity Index Measure (bottom).}}
\label{fig:training_history}
\end{figure}

To ensure reproducibility of model convergence and performance trends, the training and validation curves for Loss, PSNR, and SSIM over the 100 epochs are presented in Fig. \ref{fig:training_history}. The training and validation curves show stable convergence and no pronounced divergence between the training and validation trajectories.

\subsection{Fisher Information Calibration}
To estimate the Fisher information without corrupting the running statistics of the normalization layers, the model was placed in evaluation mode during gradient collection. We aggregated the squared gradients over a continuous calibration stream of mini-batches from the training set, applying an Exponential Moving Average (EMA) with a decay factor of $\rho = 0.9$ to stabilize the estimates. Empirical evaluation confirmed that 10 calibration batches provided stable, converged gradient variance estimates, with additional batches yielding diminishing returns.

\subsection{QUBO Solver Settings}
All QUBO problems are solved using the classical \texttt{SimulatedAnnealingSampler} from the D-Wave samplers package. The dynamic capacity search uses a single solver read per tested value of $\gamma$ to reduce search overhead, while the final QUBO solution uses 100 reads. The same optimization protocol is applied across repeated seeds.

The dynamic capacity coefficient $\gamma$ is initialized using the analytical estimate derived from the QUBO terms and refined using a binary search. The final value is then used to construct the QUBO dictionary for the final simulated-annealing solve.

\subsection{PROTES Tensor-Train Refinement Settings}
When enabled, the Tensor-Train refinement stage uses PROTES to optimize the validation PSNR of the pruning mask. The search space is binary, with one dimension per candidate pruning variable.

The final experimental configuration uses a candidate batch size of $k=300$, a maximum Tensor-Train rank of $r=10$, $k_{\mathrm{top}}=20$ elite candidates, and a learning rate of $2\times10^{-2}$. The validation objective is evaluated on a fixed validation subset to control computational cost. The QUBO mask is evaluated before refinement and is used as the initialization/reference solution for the refinement stage. 

All reported PROTES runtime and evaluation counts are measured directly during the optimization procedure.

\subsection{Qualitative Visual Results}
\begin{figure}[t!]
\centering
\includegraphics[width=\columnwidth]{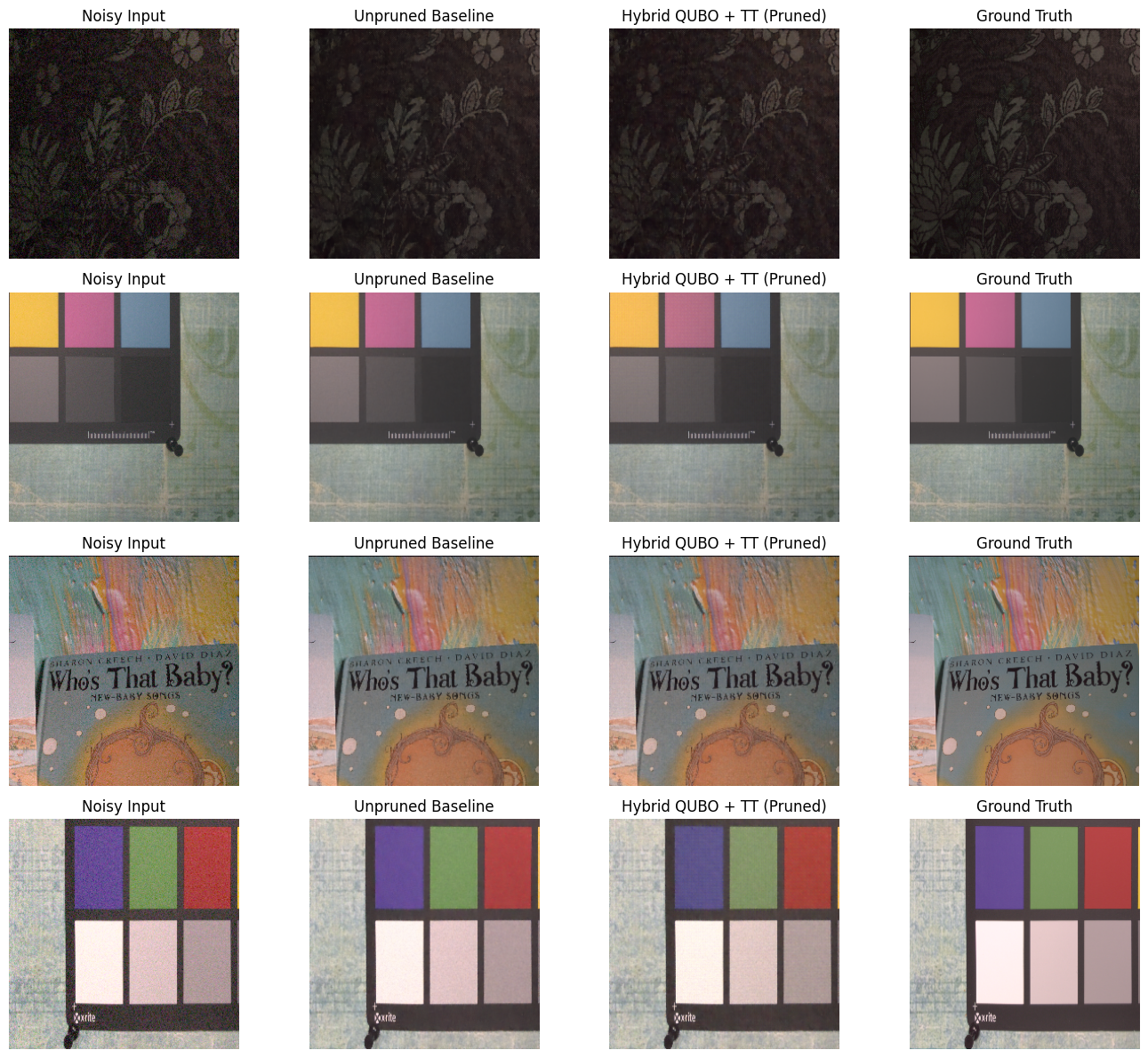}
\caption{\textbf{Qualitative comparison of image denoising results.}}
\label{fig:qualitative_results}
\end{figure}

To qualitatively assess whether pruning introduced visible degradation or artifacts, we performed visual inspections of the denoised outputs. Fig. \ref{fig:qualitative_results} provides a side-by-side comparison of the noisy input, the unpruned baseline output, our Hybrid QUBO + TT Refinement output, and the ground truth. Visual inspection shows that the pruned output preserves the main structural content of the reference image without obvious additional artifacts.

\section*{Acknowledgment}

The study was supported by the Ministry of Economic Development of the Russian Federation (agreement No. 139-10-2025-034 dd. 19.06.2025, IGK 000000C313925P4D0002)

% --- Bibliography ---
\bibliographystyle{IEEEtran}
\bibliography{ref}

% --- Biographies Placeholder ---
\begin{IEEEbiography}[{\includegraphics[width=1in,height=1.25in,clip,keepaspectratio]{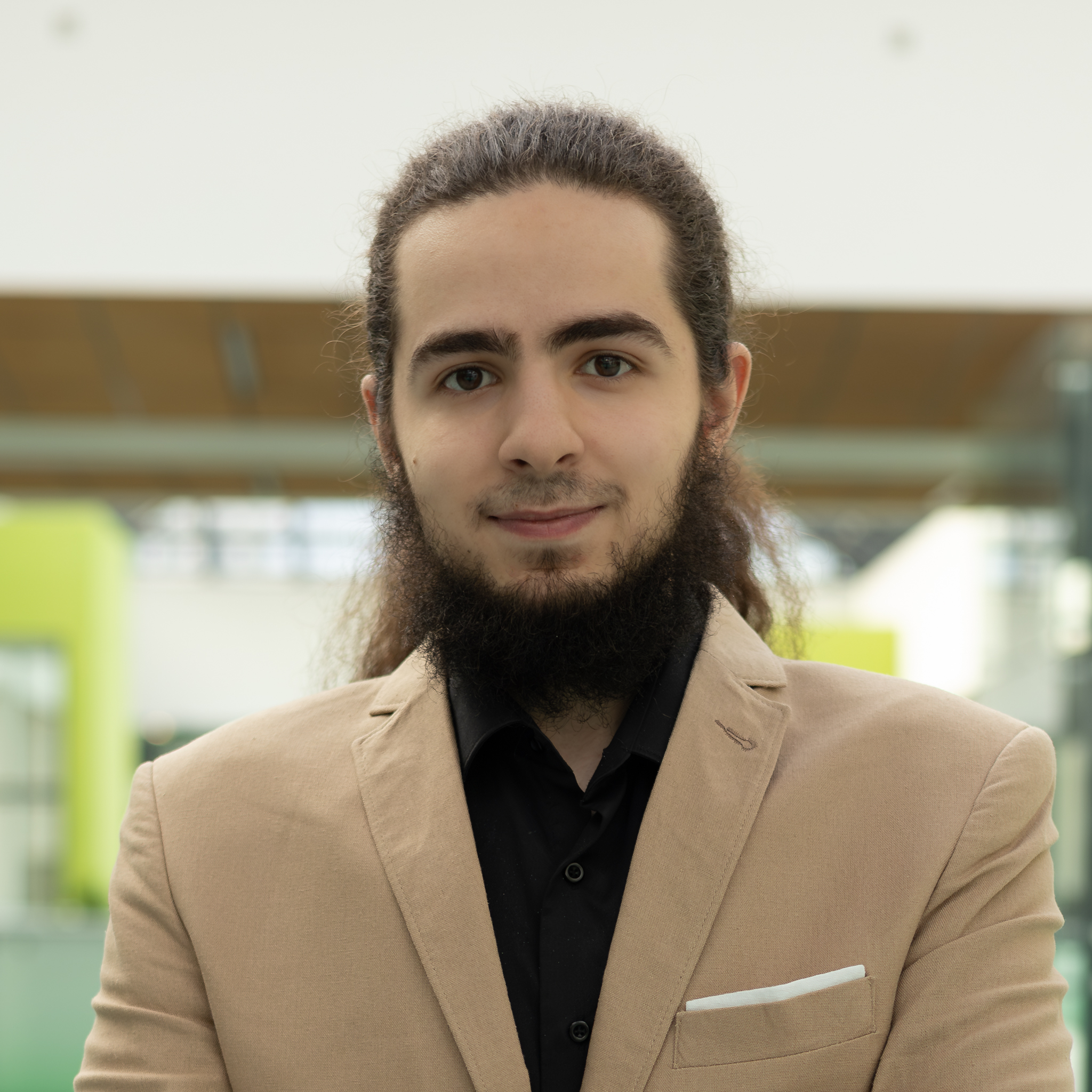}}]{Osama Orabi}
Osama Orabi received the B.S. degree in artificial intelligence and data science from Innopolis University, Innopolis, Russia, in 2026. He is a member of the Artificial Intelligence and Quantum Computing Laboratory and collaborates with the Moscow Independent Research Institute of Artificial Intelligence (MIRAI). His research focuses on combinatorial optimization for machine learning, particularly quadratic unconstrained binary optimization (QUBO) methods for neural network compression. His interests include structured pruning, mixed-precision quantization, and hybrid classical and quantum optimization.
\end{IEEEbiography}

\begin{IEEEbiography}[{\includegraphics[width=1in,height=1.25in,clip,keepaspectratio]{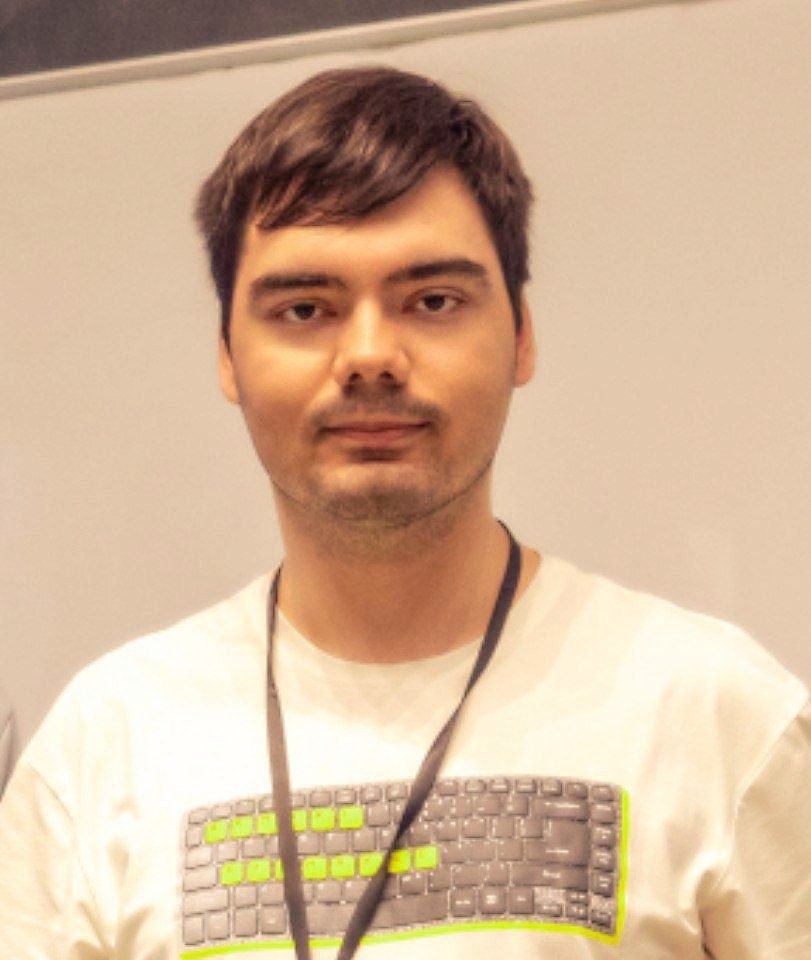}}]{Artur Zagitov}
Artur Zagitov is a PhD student at Innopolis University. His research focuses on mathematical modeling and the development of efficient methods for neural network compression, particularly pruning and quantization.
\end{IEEEbiography}

\begin{IEEEbiography}[{\includegraphics[width=1in,height=1.25in,clip,keepaspectratio]{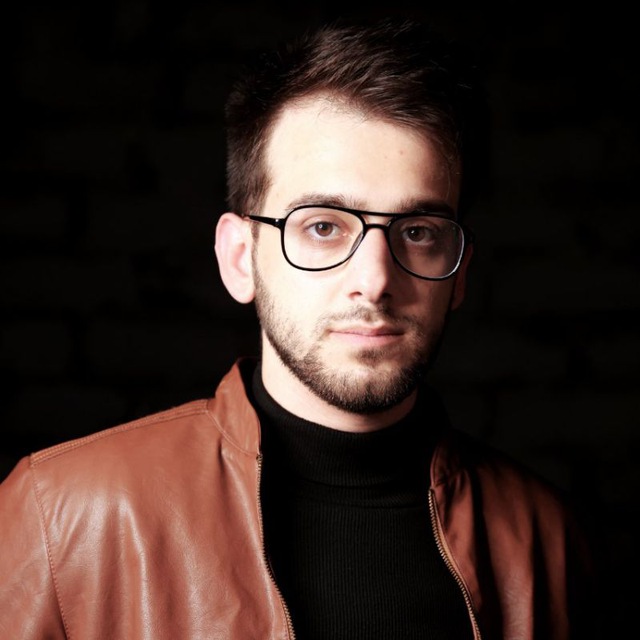}}]{Hadi Salloum}
Hadi Salloum received the B.S. degree in Artificial Intelligence and Data Science from Innopolis University, Innopolis, Russia, in 2026. He is a researcher at the Artificial Intelligence Institute at Innopolis University. His research interests focus on applying advanced artificial intelligence techniques and computational methodologies to complex real-world challenges. He has demonstrated exceptional commitment to the field of physics through his participation in prestigious international competitions, including the International Physics Olympiad (IPhO) 2021 in Lithuania and the European Physics Olympiad (EuPhO) in Estonia. His current work explores innovative applications of data science and AI, with a particular emphasis on utilizing quantum computing principles to enhance machine learning capabilities.
\end{IEEEbiography}

\begin{IEEEbiography}[{\includegraphics[width=1in,height=1.25in,clip,keepaspectratio]{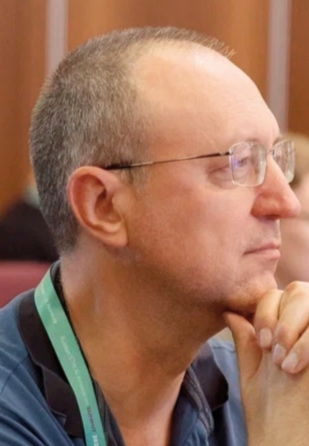}}]{Viktor A. Lobachev}
Viktor A. Lobachev is a Senior Research Fellow at the Moscow Independent Research Institute of Artificial Intelligence (MIRAI). He received his Ph.D. from the University of Rochester in 2000. His scientific areas of interest include numerical simulation and its applications in physics, artificial intelligence, and quantum computations. He has led numerous AI and mathematical modeling projects at Russian and international enterprises.
\end{IEEEbiography}

\begin{IEEEbiography}[{\includegraphics[width=1in,height=1.25in,clip,keepaspectratio]{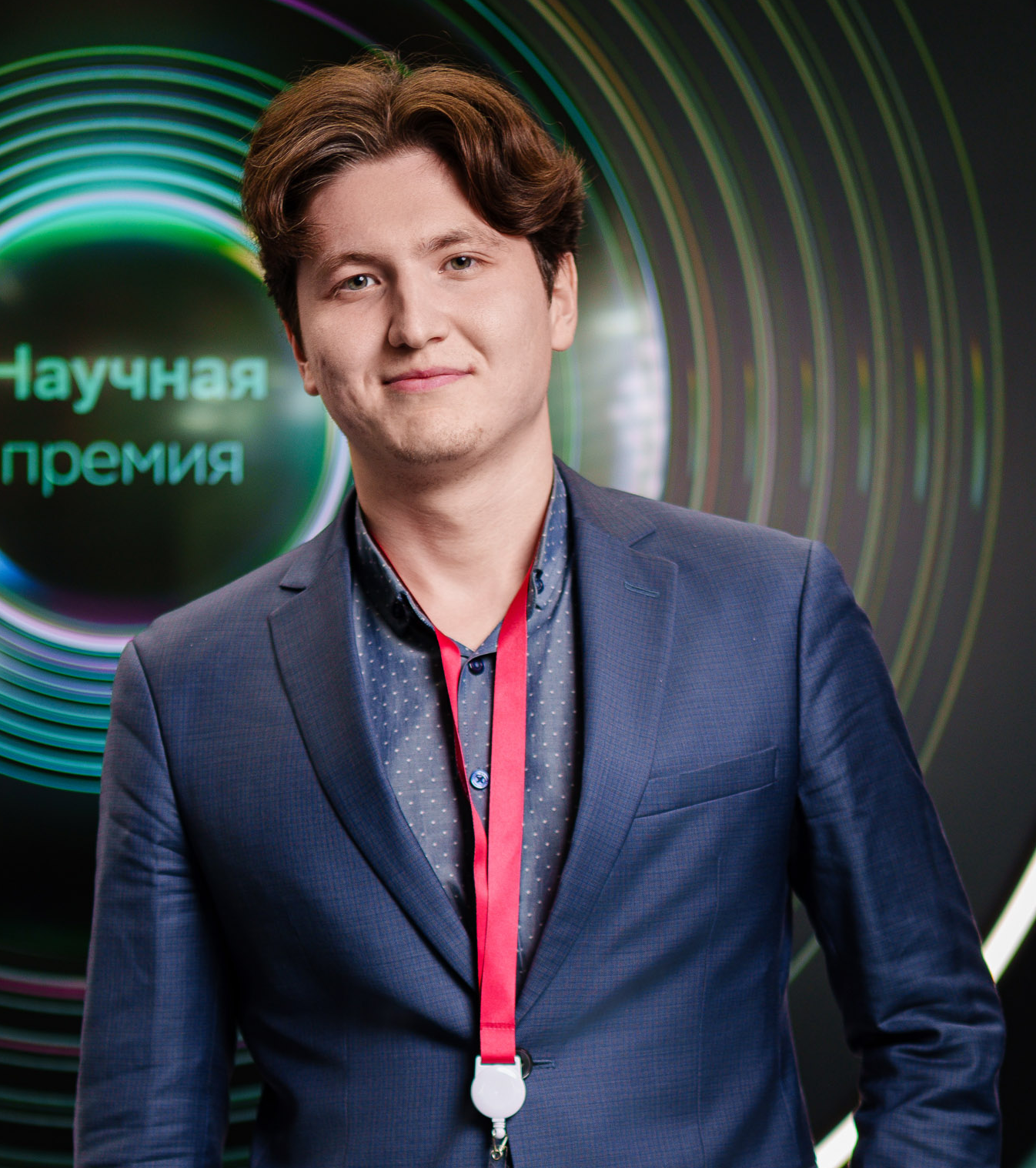}}]{Kasymkhan Khubiev}
Kasymkhan Khubiev is a researcher at the Scientific Center for Information Technologies and Artificial Intelligence, Sirius University of Science and Technology, Russia. His research focuses on applied artificial intelligence, including data-driven methods for real-world systems.
\end{IEEEbiography}

\begin{IEEEbiography}[{\includegraphics[width=1in,height=1.25in,clip,keepaspectratio]{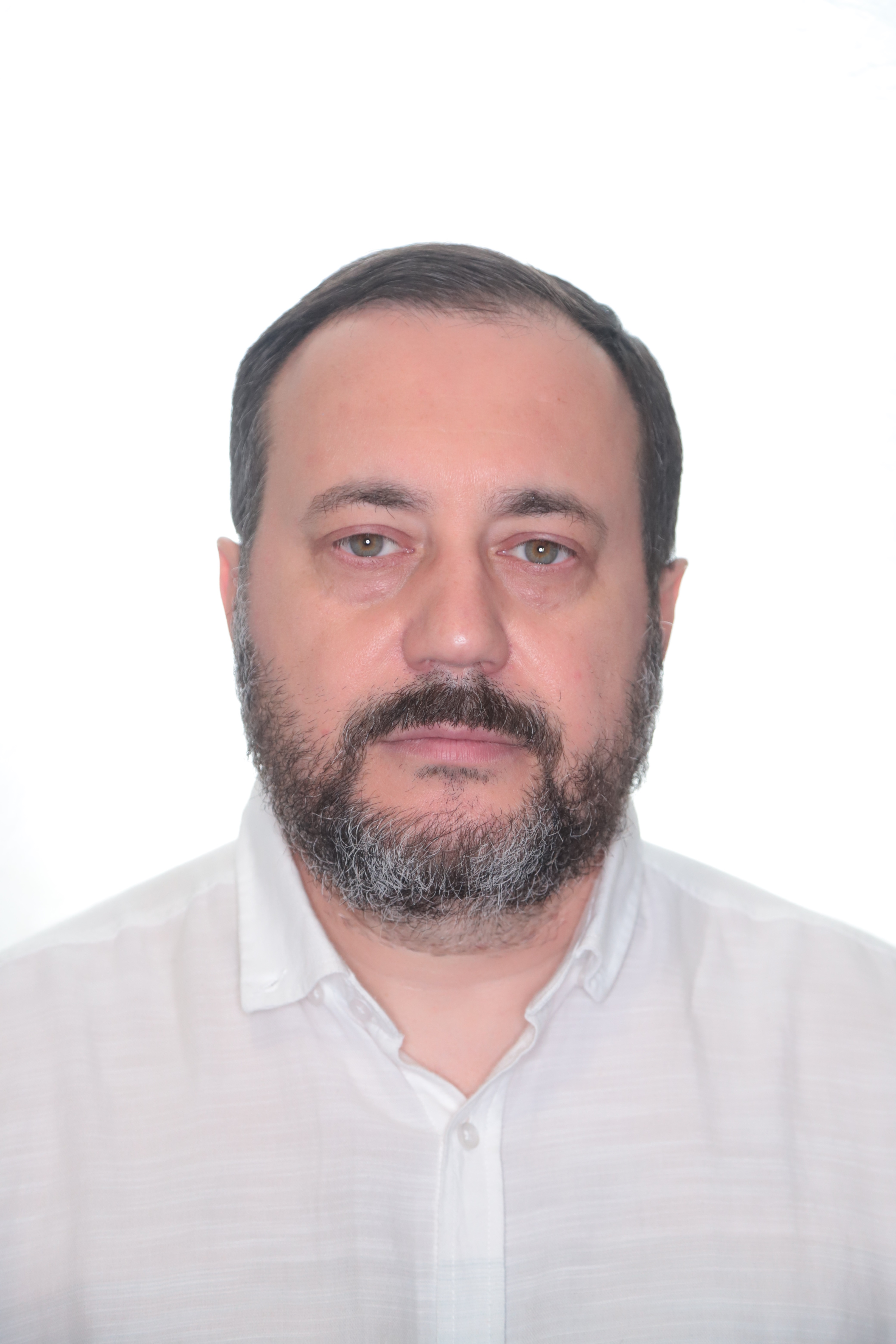}}]{Yaroslav Kholodov}
Yaroslav Kholodov is a Doctor of Sciences (Habilitation) and a Full Professor. He has been working at the Applied Mathematics department of a leading technical university in Moscow since 2001, where he has taught courses such as Computational Mathematics, Nonlinear Numerical Methods, and Traffic Modeling. Prof. Kholodov served as chair of the annual school on high-performance computing in Moscow from 2009 to 2012 and was the head of the NVIDIA CUDA Research Center from 2010 to 2015. He is the author of more than 80 scientific papers and 17 certificates of state registration for computer programs. He also serves as a reviewer for scientific journals such as Mathematics and Computer Researches and Modeling, and is the co-author of four textbooks. His research interests include data analysis, intelligent transportation systems, bioinformatics, and numerical modeling.
\end{IEEEbiography}

\EOD

\end{document}